\documentclass{IEEEtran}
\usepackage{amsmath,amssymb,amsfonts}
\usepackage{algorithmic}
\usepackage{graphicx}
\usepackage[numbers,sort&compress]{natbib}
\usepackage{textcomp}
\usepackage{stfloats}
\usepackage{xcolor}
\def\BibTeX{{\rm B\kern-.05em{\sc i\kern-.025em b}\kern-.08em
    T\kern-.1667em\lower.7ex\hbox{E}\kern-.125emX}}
\begin{document}
\title{EM-GANSim: Real-time and Accurate EM Simulation Using Conditional GANs for 3D Indoor Scenes}
\author{Ruichen Wang, and Dinesh Manocha
\thanks{R. Wang and D. Manocha are with the Department of Computer Science and the Department of Electrical and Computer Engineering, University of Maryland, College Park, MD 20742 USA (e-mail: rwang92@terpmail.umd.edu; dmanocha@umd.edu).}}

\maketitle

\begin{abstract}
 We present a novel machine-learning (ML) approach  (EM-GANSim) for real-time electromagnetic (EM) propagation that is used for wireless communication simulation in 3D indoor environments. Our approach uses a modified conditional Generative Adversarial Network (GAN) that incorporates encoded geometry and transmitter location while being guided by electromagnetic propagation priors derived from ray-tracing simulations. The overall physically-inspired learning is able to predict the power distribution in 3D scenes, which is represented using heatmaps. We evaluated our method on 15 complex 3D indoor environments, with 4 additional scenarios later included in the results, showcasing the generalizability of the model across diverse conditions. Our overall accuracy is comparable to ray tracing-based EM simulation, as evidenced by lower mean squared error values. Furthermore, our GAN-based method drastically reduces the computation time, achieving a 5X speedup on complex benchmarks. In practice, it can compute the signal strength in a few milliseconds on any location in 3D indoor environments. We also present a large dataset of 3D models and EM ray tracing-simulated heatmaps. To the best of our knowledge, EM-GANSim is the first real-time algorithm for EM simulation in complex 3D indoor environments. We plan to release the code and the dataset. 
\end{abstract}


\section{Introduction}
\label{sec:introduction}
\IEEEPARstart{E}{lectromagnetic} (EM) waves, characterized by the oscillation of electric and magnetic fields, are central to technologies such as visible light, microwave ovens, and wireless communication systems, including Wi-Fi and 5G. Maxwell's equations \citep{maxwell1873treatise} describe the interaction and propagation of these electric and magnetic fields through space, forming the theoretical foundation for understanding EM wave behavior, including reflection, refraction, diffraction, and scattering. These principles are critical when modeling wave propagation in complex environments \citep{obaidat2003simulation, d2018linelab}, such as indoor spaces, where multiple interactions with obstacles and materials occur.

In the field of EM simulation, various methods are employed to understand wave propagation and interaction with media. Path loss and attenuation play crucial roles in these simulations, measuring how much signal power diminishes over distance, due to obstacles or the medium itself. Ray tracing is a widely-used technique for simulating wave interactions with surfaces \citep{bertoni1994uhf, seidel1992914}, as it balances computational efficiency and accuracy. It simulates rays, as narrow beams of EM energy, traveling in straight lines and accounting for key phenomena like reflection and diffraction. The method's efficiency makes it popular for 5G network planning \citep{hsiao2017ray}, vehicular communications \citep{wang2022dynamic}, electromagnetic characterization \citep{egea2021opal}, and ground-penetrating radar \citep{zhang2006ray}. Other methods such as wave-based methods that numerically solve Maxwell's equations can provide more accurate results, capturing complex wave behavior such as diffraction and scattering more precisely. However, these methods are often too computationally intensive for real-time or large-scale applications ~\citep{coifman1993fast, taflove2005computational},

Despite its advantages, current EM simulation systems, particularly those based on ray tracing, have limitations in terms of handling dynamic scenes or complex environments. Ray tracing relies on modeling rays, i.e., narrow beams of EM energy, that travel in straight lines until they encounter an object, tracing their paths from a source and modeling interactions like reflections and diffractions~\citep{mckown1991ray}. The simulation accuracy depends on detailed environmental models and material properties, making it computationally intensive, and needs significant processing power to simulate the numerous potential ray paths in complex environments. For dynamic scenes and detailed indoor environments, the need to continually update the models and recompute new paths in real-time is a major challenge. Indoor simulations are particularly difficult due to the complexity and density of the obstacles, which further increases the computational load, making current methods inefficient for applications requiring quick responses, such as 5G network planning,  where higher frequencies and more complex environments are used~\citep{wang20206g,rappaport2013millimeter}. 

Innovative solutions, such as integrating generative adversarial networks (GANs) into EM simulations, are being explored to address these limitations. GAN models include considerations for path loss, reflection, and diffraction in their loss functions, and they also account for material properties and multipath propagation, thereby improving simulation accuracy and heatmap generation for real-world applications.

\textbf{Main Results:} We present a novel GAN-based prediction scheme for real-time EM simulation in 3D indoor scenes ~\citep{wangthesis}. Our formulation uses a physically-inspired generator to predict wireless signal received power heatmaps and ensures high accuracy by incorporating detailed signal propagation mechanisms such as direct propagation, reflection, and diffraction. These physical constraints are embedded within the GAN's loss function to ensure that the generated data adheres to the principles of electromagnetic wave propagation. We use ray tracing techniques to model how signals propagate through an environment, considering reflections off surfaces and diffractions around the obstacles~\citep{sangkusolwong2017indoor}. We evaluate these physical interactions using EM propagation models and the uniform theory of diffraction (UTD)~\citep{kanatas1997utd} to predict the path loss for indoor environments accurately. The primary evaluation of our method is conducted across 15 distinct indoor scenes, and additional scenes are later incorporated to further demonstrate the versatility of our approach in various environments. Our approach not only improves the reliability of the heatmap predictions but also enhances the robustness and convergence of the GAN during training. Our main contributions include:

\begin{itemize}
    \item {\em Accurate Power Distributions}: By employing conditional Generative Adversarial Networks (cGANs) and utilizing the strengths of physics-inspired learning, our approach can predict accurate power distributions in 3D indoor environments. 
    \item {\em Real-Time Performance}: We highlight the performance on 15 complex 3D indoor benchmarks. Our approach significantly reduces the computational time needed for simulations compared to prior methods based on ray tracing. Our GAN models streamline the simulation process, achieving 5X faster running time on entire power map generation for various-sized indoor models. Additionally, it enables real-time simulation for individual data points in just a few milliseconds.
    \item {\em Dataset}: We present a large, comprehensive dataset featuring varied indoor scenarios (2K+ models) and simulated heatmaps (more than 64M) to train our model. This dataset ensures robust and generalized model performance across diverse conditions and is used for training and testing.  
\end{itemize}

Unlike prior machine-learning approaches to pathloss prediction that typically regress scalar values, local features, or point-wise radio-map values from geometry or measurements, EM-GANSim predicts the full spatial distribution of received power using conditional generation. DeepRay uses encoder-decoder networks with ray-tracing data, PL-GAN applies GANs to path-loss prediction, and the First Pathloss Radio Map Prediction Challenge studies data-driven radio-map prediction. EM-GANSim differs from these methods by combining full-field heatmap generation, 3D geometry and transmitter conditioning, and DCEM-derived propagation masks used as physics-guided supervision. This combination helps preserve physically meaningful spatial patterns, including reflection- and diffraction-consistent regions, while retaining real-time inference speed \citep{yapar2024overview,bakirtzis2022deepray,marey2022pl,yun2015ray}.

\section{EM Propagation Background}

EM wave propagation determines how wireless signals travel through indoor environments and how they are modified by scene geometry and materials. Wireless propagation is ultimately governed by Maxwell's equations, but in this work we rely on ray-based approximations that are more directly relevant to the proposed learning framework. We summarize here only the propagation concepts that are directly used in our model design and loss construction; readers are referred to broader treatments of electromagnetic propagation and ray tracing for full derivations and additional physical effects \citep{b24,b25,b26}.

\subsection{Path Loss and Wave Interaction}

As an EM wave propagates, its power decays with distance and is further altered by environmental interactions. A common reference model is the close-in path loss expression
\begin{align}
PL^{CI}(f, d)[dB] &= FSPL(f, d=1m)[dB] \\
                  &\quad + 10 \log_{10}(d)[dB] + AT[dB], \notag
\end{align}
where \(f\) is the carrier frequency, \(d\) is the propagation distance, and \(AT\) denotes atmospheric attenuation \citep{okoro2021modeling}.

In indoor settings, the received power is shaped not only by distance-dependent attenuation but also by wave interactions with the environment. Reflection from walls, windows, and furniture modifies signal strength and can create constructive or destructive interference \citep{b1}. Diffraction allows waves to bend around edges and corners, which is important in obstructed layouts \citep{b24,b25}. Scattering redistributes energy from rough or irregular surfaces and can contribute both specular and non-specular components depending on the surface characteristics \citep{b26,yun2015ray}. These mechanisms jointly produce the structured spatial variations observed in received-power maps.

\subsection{Material Effects}

Material properties such as permittivity and conductivity influence reflection, transmission, absorption, and attenuation. Consequently, common indoor materials such as concrete, glass, and wood lead to different propagation behaviors, making material-aware scene representation important for realistic received-power prediction \citep{b24,b25,b26}.

\subsection{Ray-Tracing-Based Propagation Modeling}

Ray tracing approximates EM propagation by modeling direct, reflected, transmitted, diffracted, and scattered paths through the scene. It provides a practical balance between physical fidelity and computational efficiency and is widely used for wireless planning and indoor propagation analysis \citep{wang2022dynamic}. In this work, ray tracing serves two purposes: it provides the reference heatmaps used for training and evaluation, and it supplies propagation-aware spatial structure used in the proposed physics-guided losses.

\section{Prior Work}

Recent efforts in integrating ML with EM ray tracing and wireless communication systems have highlighted the potential of ML in enhancing wireless communication technologies in various ways. DeepRay~\citep{bakirtzis2022deepray} uses a data-driven approach that integrates a ray-tracing simulator with deep learning models, specifically convolutional encoder-decoders such as U-Net and SDU-Net, enhancing indoor radio propagation modeling for accurate signal strength prediction in various indoor environments. The model is able to learn from multiple environments and predict unknown geometries with high accuracy. WAIR-D~\citep{huangfu2022wair} introduces a comprehensive dataset supporting AI-based wireless research, emphasizing the creation of realistic simulation environments for enhanced model generalization and facilitating fine-tuning for specific scenarios using real-world map data. Huang et al.~\citep{huang2021collaborating} integrate ray tracing and an autoencoding-translation neural network to perform 3-D sound-speed inversion, improving efficiency and accuracy in underwater acoustic applications.  Yin et al.~\citep{yin2022millimeter} investigate the use of millimeter wave (mmWave) wireless signals in assisting robot navigation and employ a learning-based classifier for link state classification to enhance robotic movement and decision-making in complex environments. There are other methods that combine deep reinforcement learning with enhanced ray tracing for antenna tilt optimization and those leveraging 5G MIMO data for beam selection using deep learning techniques to improve cellular network performance through efficient geospatial data processing and precise signal optimization~\citep{zhu2022ai, wang2023cellular}.

ML techniques have also been used to predict the received power in complex indoor and urban environments ~\citep{yun2015ray}. Traditional methods like regression models, decision trees, and support vector machines have been used to model the propagation characteristics of electromagnetic fields. The performance of these methods has been improved by adapting to data from specific environments, thereby enhancing prediction accuracy for both line-of-sight (LoS) and non-line-of-sight (NLoS) conditions~\citep{filosa2016ray,dong2020application, williams2015ray}. 

Despite their advancements, traditional machine learning methods, such as regression models and support vector machines, as well as conventional simulation techniques like ray tracing, face significant limitations in capturing the highly nonlinear interactions and multipath effects characteristic of indoor and urban EM propagation. These challenges become more pronounced when modeling dynamic environmental changes, such as moving objects and varying channel conditions, which conventional approaches typically struggle to address effectively~\citep{marey2022pl}. In contrast, our choice of modified conditional Generative Adversarial Networks (cGANs)~\citep{creswell2018generative} offers unique advantages in generating realistic synthetic data and accurately predicting heatmaps, making them particularly well-suited for wireless communication network design in dynamic environments.

\section{Methodology}
\subsection{Overview}
In this section, we present our novel approach for augmenting EM ray tracing techniques with a modified cGAN. Our goal is to design a simulator for 3D indoor scenes, the accuracy of which is similar to that of EM ray tracers but is significantly faster for real-time or dynamic scenarios.  Figure~\ref{fig1} shows the overall architecture of our network:
\begin{figure}[htbp]
\centerline{\includegraphics[scale=0.3]{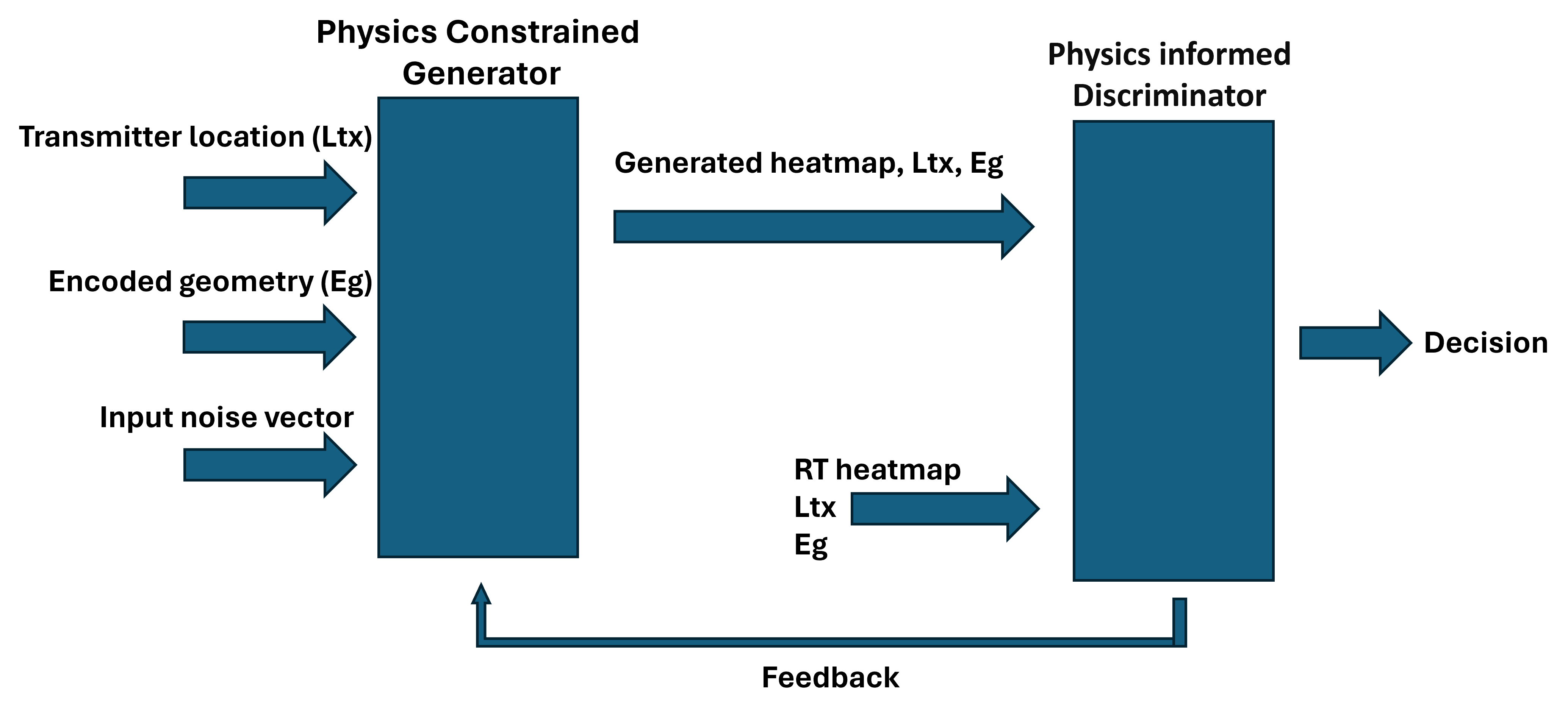}}
\caption{Overall architecture of our cGAN training process. The Generator (G) takes encoded 3D geometry, transmitter location, and a noise vector to output simulated heatmaps. The Discriminator (D) evaluates both the real heatmap from a ray-tracing simulator DCEM and the generated heatmap from G and makes 0/1 decisions.}
\label{fig1}
\end{figure}

Our network's formulation can be described as follows:
\begin{equation}
P_{\text{r}} = f_{\text{cGAN}}(E_{\text{g}}, L_{\text{tx}}, z)
\end{equation}
where
\begin{itemize}
    \item $P_r$ is the predicted received-power heatmap over the target 2D evaluation plane at a fixed height level.
    \item $f_{\text{cGAN}}$ denotes the function computed using the conditional Generative Adversarial Network. It models the complex relationship between the indoor environment's geometry, material properties, the transmitter's location, and the resulting EM signal heatmap.
    \item $E_{\text{g}}$ represents the encoded geometry information of the indoor environment. It encapsulates details such as the spatial layout in 2D with height information and material properties, which are used for accurate EM propagation modeling.
    \item $L_{\text{tx}}$ refers to the precise location of the transmitter within the environment. The transmitter's position, in conjunction with the environment's geometry, significantly impacts EM wave propagation and the distribution of received power.
    \item $z$ is a zero-mean unit-variance Gaussian noise vector concatenated with the conditioning inputs to make training less sensitive to symmetry artifacts and mode collapse.
\end{itemize}

\textbf{Geometry Encoding and Noise:}
The encoded geometry $E_g$ is obtained by rasterizing each 3D indoor scene into a top-view grid of resolution $256 \times 256$. The resulting input tensor has size $256 \times 256 \times C$, with
\[
C = C_{\mathrm{mat}} + 1,
\]
where $C_{\mathrm{mat}}$ denotes the number of material-related channels and the final channel stores normalized height values. This representation preserves the floor-plan layout while retaining material and vertical information needed for propagation modeling. In particular, walls, doors, and windows are mapped to discrete grid locations, and the associated material encoding allows the network to distinguish propagation behavior across different surface types. Material-dependent propagation effects are further incorporated through the physics-guided supervision derived from ray-tracing simulations, which implicitly capture reflection, absorption, and diffraction behaviors associated with different materials. In this formulation, the network does not require explicit parameterization of electromagnetic properties (e.g., permittivity or conductivity), but instead learns their effective impact on field distributions under physically consistent training signals.

The noise vector $z$ is sampled from a zero-mean unit-variance Gaussian distribution and concatenated with the conditioning inputs to the generator. This stochastic input makes training less sensitive to symmetry artifacts and mode collapse.

\subsection{Modified Conditional GAN}

We use a conditional GAN because the target output is a spatially structured received-power map rather than a single scalar prediction. Compared with conventional regression models, the cGAN formulation better captures the high-dimensional distribution of indoor EM fields conditioned on scene geometry and transmitter location. This is particularly important in indoor environments, where received power depends on highly nonlinear interactions among line-of-sight propagation, reflections, diffraction, and material-dependent attenuation.

Accordingly, our generator is conditioned on the encoded geometry $E_g$, the transmitter location $L_{tx}$, and noise $z$, and produces a received-power heatmap for the target scene. The discriminator is trained to distinguish generated heatmaps from ray-traced reference maps, while the generator is additionally regularized using physics-based losses derived from propagation priors. This formulation allows the model to retain the efficiency of a learned predictor while enforcing physically meaningful structure in the generated field distributions.

The generator outputs a single aggregate received-power heatmap $\hat{P}$. To incorporate propagation semantics, we construct three DCEM-derived spatial masks associated with direct, reflected, and diffracted regions. The propagation masks $M_k$ are derived directly from ray-tracing outputs, where each spatial region is labeled according to the dominant propagation mechanism (e.g., line-of-sight, reflection, or diffraction) based on ray interaction classification. For each mask, we compare the masked response of the generated heatmap with that of the ray-traced reference map. In this way, the physical losses act on region-wise propagation structure rather than requiring the network to predict separate component maps.

\begin{equation}
\mathcal{L}_{\text{cGAN}}^{\text{G}} = \mathbb{E}_{E_{\text{g}}, L_{\text{tx}}, z}[\log(1 - D(E_{\text{g}}, L_{\text{tx}}, G(E_{\text{g}}, L_{\text{tx}}, z)))]
\end{equation}
This equation represents the adversarial loss for the generator $G$ in the cGAN and aims to minimize the ability of discriminator $D$ to distinguish generated heatmaps from real ones.
The Mean Squared Error (MSE) loss measures the discrepancy between the real received power and the power predicted by the generator, given below:
\begin{equation}
\mathcal{L}_{\text{MSE}} = \mathbb{E}_{E_{\text{g}}, L_{\text{tx}}, P_{\text{r}}}[\|P_{\text{r}} - G(E_{\text{g}}, L_{\text{tx}}, z)\|_2^2]
\end{equation}

\subsubsection{Generator}
Our generator uses a series of convolutional neural network (CNN) layers designed to capture the intricate spatial relationships within indoor environments. Special attention is given to encoding the geometry information effectively, allowing the model to understand how different materials and layouts affect signal propagation. We also incorporate physical constraints into the objective function, ensuring that the generated samples adhere to the fundamental principles of electromagnetic wave propagation. The generator objective function is given as:
\begin{equation}
\begin{split}
\mathcal{L}_{\text{G}} = 
& \mathcal{L}_{\text{cGAN}}^{\text{G}} 
  + \lambda \mathcal{L}_{\text{MSE}} 
  + \mu \mathcal{L}_{\text{phy}}.
\end{split}
\end{equation}

This equation combines the cGAN loss with the MSE loss balanced by a weighting factor $\lambda$. Additionally, $\mathcal{L}_{\text{phy}}$ represents the physical constraints loss, and $\mu$ is a weighting factor that balances the importance of the physical constraints in the overall objective function. The physical constraints loss $\mathcal{L}_{\text{phy}}$ is defined as
\begin{equation}
\mathcal{L}_{\text{phy}}
=
\alpha\,\mathcal{L}_{\text{direct}}
+
\beta\,\mathcal{L}_{\text{reflection}}
+
\gamma\,\mathcal{L}_{\text{diffraction}},
\end{equation}
where $\alpha$, $\beta$, and $\gamma$ weight the contributions of direct, reflected, and diffracted propagation, respectively.

Let
\begin{equation}
S_k(P)=\sum_{x,y} M_k(x,y)P(x,y),
\end{equation}
where $P$ denotes either the generated heatmap $\hat{P}$ or the reference heatmap $P^{\ast}$, and $k \in \{\text{direct}, \text{reflection}, \text{diffraction}\}$.

The component-wise physical losses are then defined as
\begin{equation}
\mathcal{L}_{\text{direct}}
=
\left(
S_{\text{direct}}(\hat{P})-S_{\text{direct}}(P^{\ast})
\right)^2,
\end{equation}

\begin{equation}
\mathcal{L}_{\text{reflection}}
=
\left(
S_{\text{reflection}}(\hat{P})-S_{\text{reflection}}(P^{\ast})
\right)^2,
\end{equation}

\begin{equation}
\mathcal{L}_{\text{diffraction}}
=
\left(
S_{\text{diffraction}}(\hat{P})-S_{\text{diffraction}}(P^{\ast})
\right)^2.
\end{equation}

In this formulation, the generator predicts a single aggregate received-power heatmap rather than separate multipath-component maps. 
The three physical losses therefore act on region-wise responses defined by DCEM-derived propagation masks, encouraging the generated heatmap to preserve the relative contributions of the dominant propagation mechanisms without requiring explicit per-component output channels.

We emphasize that the proposed model does not explicitly predict per-path components (e.g., individual direct, reflected, or diffracted rays). Instead, the quantities $S_k(\cdot)$ represent \emph{region-wise aggregated responses} over propagation masks derived from ray-tracing simulations. This formulation allows the network to capture the relative contribution of different propagation mechanisms while maintaining a single physically consistent output field. As a result, the model focuses on accurately predicting the aggregate received-power distribution, which is the primary quantity used in practical wireless simulation and network planning tasks.

\subsubsection{Discriminator}
Our discriminator is also based on CNNs, with the addition of condition layers that incorporate the geometry information. This setup ensures that the discrimination process considers not just the accuracy of the heatmaps but also their consistency with the input geometry. This consistency refers to a check of the alignment of predicted signal strengths with the expected patterns based on EM propagation theory discussed earlier in the generator, such as maintaining the correct spatial distribution and intensity of signals influenced by environmental factors and material properties. The Discriminator Objective Function is given as:
\begin{equation}    
\begin{aligned}
\mathcal{L}_{\text{cGAN}}^{\text{D}} = &-\mathbb{E}_{E_{\text{g}}, L_{\text{tx}}, P_{\text{r}}}[\log D(E_{\text{g}}, L_{\text{tx}}, P_{\text{r}})] \\
&-\mathbb{E}_{E_{\text{g}}, L_{\text{tx}}, z}[\log(1 - D(E_{\text{g}}, L_{\text{tx}}, G(E_{\text{g}}, L_{\text{tx}}, z)))]
\end{aligned}
\end{equation}

This function models the discriminator's objective, which seeks to identify real and generated heatmaps correctly, thus ensuring that the generated data is accurate.

\subsection{Training}
Training of the modified cGAN is performed using a loss function that balances the fidelity of the generated heatmaps as a function of the input geometric conditions. The training process is carefully monitored to prevent mode collapse and ensure a diverse set of realistic outputs. We carefully balanced the weights of physical regularizations by conducting experiments  with a range of weighting factors for each physical constraint (direct propagation,  reflection, and diffraction losses) to evaluate their impact on the stability of training and 

the accuracy of predictions. Specifically, we started with baseline weights informed by  theoretical insights from electromagnetic wave propagation and iteratively adjusted these values using a grid search methodology to optimize both convergence and output fidelity. 
Each configuration was evaluated based on metrics such as MSE and training loss dynamics to ensure stability. These constraints improve prediction accuracy while avoiding excessive penalization, which could destabilize the generator. 

\begin{figure}[htbp]
\centerline{\includegraphics[scale=0.3]{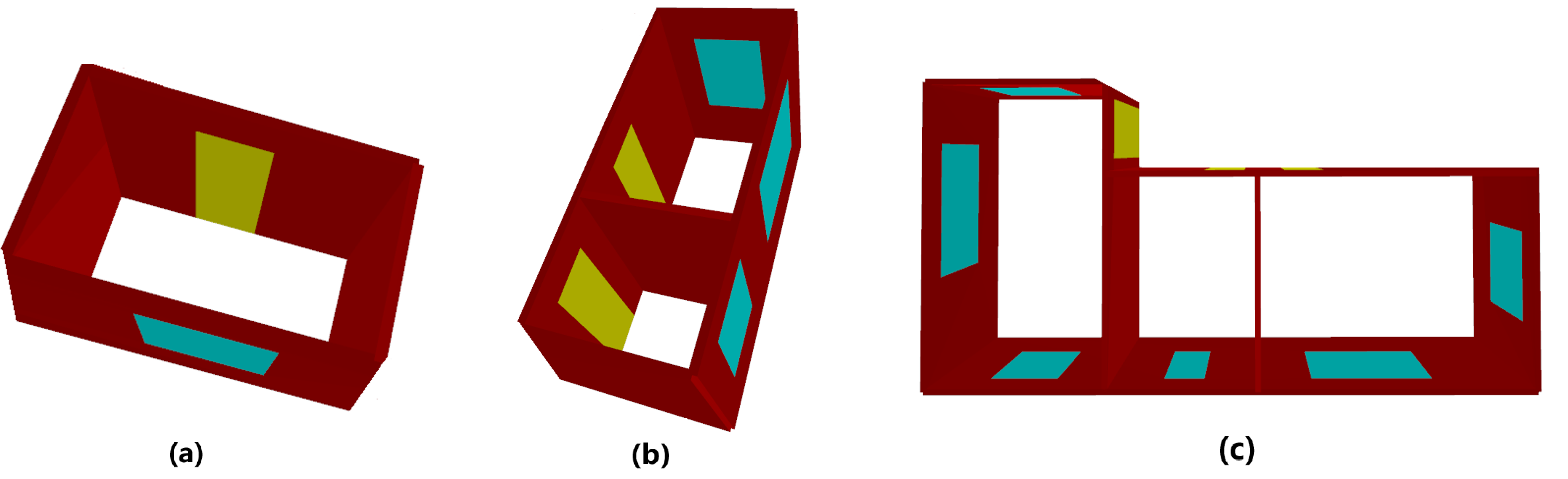}}
\caption{Representative examples of the indoor-scene geometry used in our experiments. We show three canonical layouts: (a) Single-room setup with minimal furniture. (b) Multi-room configuration with complex wall structures. (c) Multi-room layout with varied dimensions and partitions,  drawn from our full dataset (> 2 000 models). These scenes demonstrate the diversity of layouts the ML model must interpret for accurate EM ray tracing simulation. The red represents concrete walls, the blue represents glass, and the yellow represents wooden doors. These images serve purely as illustrative samples; quantitative evaluations are reported on the 15 (baseline) + 4 (additional) benchmarks detailed in Tables II–V and Figs 4–9.}
\label{fig4}
\end{figure}

\begin{figure*}[htbp]
\centering
\includegraphics[width=\textwidth]{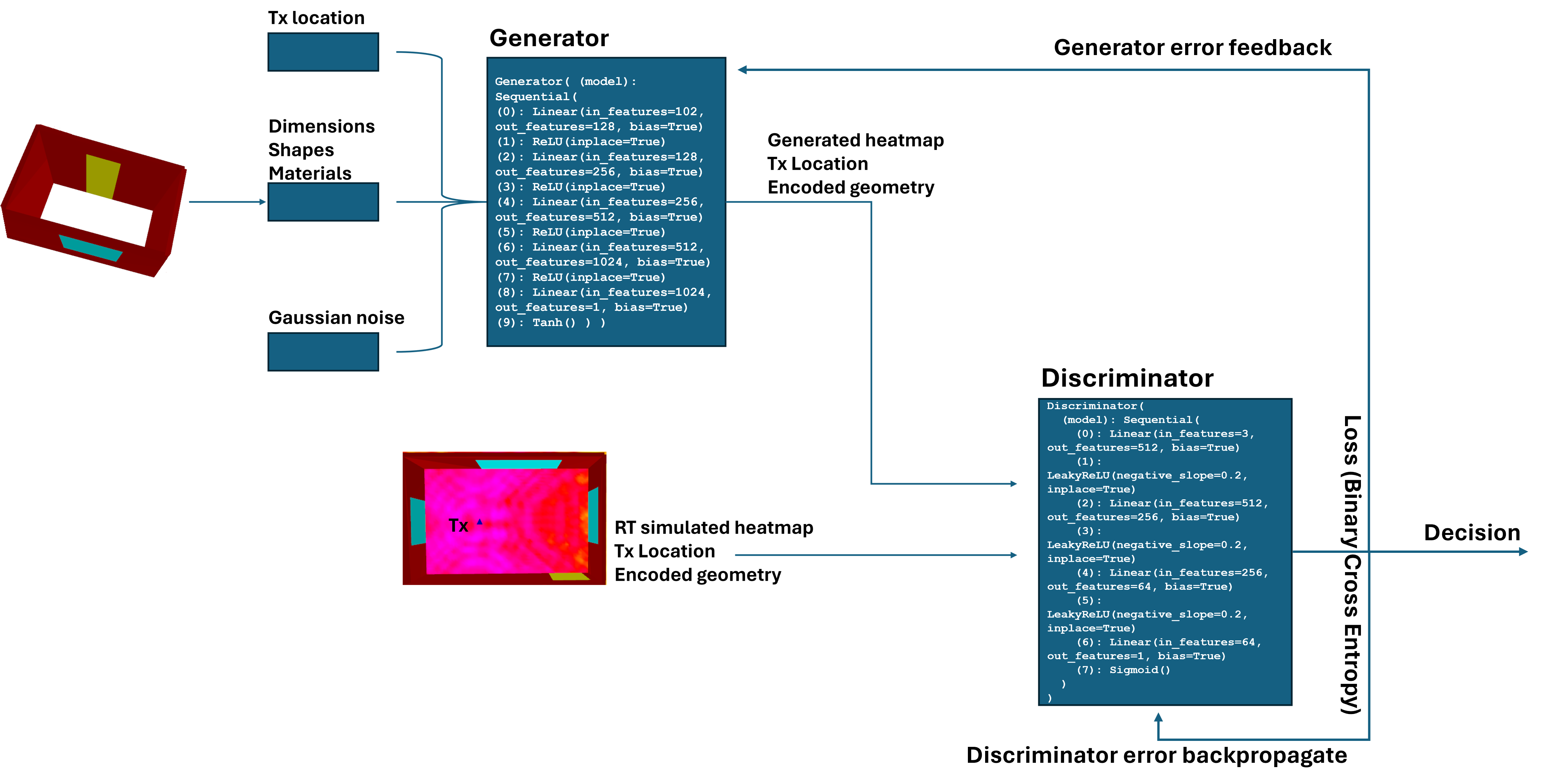}
\caption{A detailed flowchart illustrating the GAN training pipeline. The diagram summarizes four main stages: data preparation and geometry encoding, design of the generator architecture, design of the discriminator architecture, and optimization with adversarial and physics‑guided losses. This concise overview highlights the key components of the training process in a structured and professional manner.}
\label{fig5}
\end{figure*}

The proposed method is implemented using PyTorch \citep{paszke2019pytorch} and uses a GPU for efficient model training and inference. For ease of access, we utilize Google Colab, which provides free GPU resources to facilitate the training process. The primary software and dependencies include Python 3 or higher and essential libraries such as NumPy, SciPy, and Matplotlib for data handling and visualization. Our training process on Google Colab takes approximately two days to complete. We optimize the training using hyperparameters such as the learning rate, batch size, and latent space dimensions, which are crucial for achieving the desired model performance and accuracy. 
A detailed flowchart of the GAN training process and implementation details is presented below Fig.~\ref{fig5} and we plan to release our code at the time of publication. 
Sample 3D renderings of indoor environments used in the training set are shown in Fig.~\ref{fig4}, which displays three representative indoor layouts chosen solely to visualize the geometric diversity of the dataset. The complete benchmark set used for quantitative evaluation comprises 15 baseline scenes (Figs 4, 5, 8, 9) and 4 additional scenes (Fig. 7).

\section{Implementation and Performance}
We discuss the implementation of our approach and the main issues in terms of obtaining good performance.
\subsection{Data Adequacy and Quality}
Given the complexity of indoor wireless systems, the cGAN would require extensive and high-quality training data that accurately represents the vast array of environmental factors affecting signal propagation. In our process, we generate geometry and power prediction data from WinProp and the DCEM simulator to ensure the diversity and volume of training data, representing different scenarios with high quality.

In our experiments, we use a dataset of more than 2,000 indoor scenes. We allocate 1,600 layouts (80\%) for training and 400 layouts (20\%) for validation, with no overlap between the two sets. The 15 benchmark scenes used in the quantitative evaluation form an additional held-out test set, and four extra scenes are used only for qualitative generalization analyses. This split is designed to evaluate whether EM-GANSim generalizes to scene layouts that are not used during training.

\subsection{Hyperparameter Tuning}
cGANs are notoriously difficult to train and often sensitive to the choice of hyperparameters, which would require extensive experimentation to fine-tune. We employed a structured and systematic approach to hyper-parameter tuning and training complexity management. Specifically, we conducted a grid search coupled with sensitivity analysis to identify optimal hyper-parameter ranges, performing a finer search within promising intervals. Hyper-parameters were validated against a held-out dataset, ensuring model generalizability and mitigating overfitting. GAN-specific parameters, including learning rates, optimizer beta values, and adversarial loss weights, were initially set according to best practices and iteratively refined based on observed stability and performance trends. Concurrently, our training strategy involved starting with simplified versions of the environment to train the cGAN, progressively increasing complexity. This incremental approach enabled the model to grasp fundamental principles before addressing intricate scenarios, effectively avoiding convergence issues and ensuring stable, realistic simulations of EM ray tracing.
\begin{table}[h]
\centering
\caption{Hyperparameters used in GAN Training}
\label{table:Hyperparameters}
\resizebox{0.8\columnwidth}{!}{
\begin{tabular}{|l|c|}
\hline
\textbf{Hyperparameter} & \textbf{Value/Type} \\
\hline
Learning Rate & 0.0002 \\
\hline
Batch Size & 128 \\
\hline
Noise Type & Gaussian \\
\hline
Loss Function & Binary Cross Entropy \\
\hline
\end{tabular}
}
\end{table}

\subsection{Mode Collapse}
A common issue with cGANs occurs when the generator starts producing a limited range of outputs, which in the case of EM ray tracing could lead to underrepresentation of the solution space. In our work, we mitigate mode collapse and promote robust feature learning through several key strategies. First, we employ a large, diverse training dataset of over 2,000 indoor scenes, ensuring the generator is exposed to a wide range of scenarios and enhancing its generalization capability. Second, we utilize incremental training, beginning with simpler environments and gradually introducing complexity, enabling the generator to master fundamental signal propagation patterns before progressing to more intricate layouts. Third, we incorporate a noise vector into the generator’s input to further enhance feature variability and output diversity. Additionally, our conditional GAN framework ensures the relevance of generated outputs, while adaptive and dynamic learning rate schedules effectively balance generator-discriminator dynamics. Collectively, these strategies significantly enhance the model’s ability to produce diverse and accurate EM propagation heatmaps.

\section{Results and Evaluations}

\label{sec:results}
This section presents the evaluation of the proposed methodology in terms of accuracy enhancement and efficiency improvement in ray tracing simulations in 3D indoor environments. We have conducted a comparison with WinProp \citep{jakobus2018recent}, which is widely recognized as a state-of-the-art solution in EM simulation, as shown in these and more papers \citep{vaganova2023developing, wang2023dynamic, haron2021performance, gomez2023design}. We emphasize that the goal of this work is fast and accurate field-level prediction for practical wireless simulation tasks, rather than explicit decomposition of individual propagation paths. This design choice reflects the requirements of real-world applications, where aggregate field strength is the primary quantity of interest. 

We use DCEM and WinProp for two complementary purposes. DCEM provides the ray-traced reference maps used during training and for primary fidelity assessment, since the proposed physics-based losses are derived from DCEM-generated propagation structure. WinProp is used as an additional independent ray-tracing baseline to evaluate cross-simulator consistency. Therefore, comparisons against DCEM measure agreement with the reference simulator used to supervise training, whereas comparisons against WinProp indicate whether the learned model also remains consistent with a separate industry-standard implementation.

We show evaluations in 15 indoor scenes: Scenes 1-15. Detailed specifications of scenes are included in Table \ref{table:detailed}. On average, the running time of EM-GANSim across the evaluated indoor scenes is approximately 1 ms per data point. However, the models with complex layouts tend to require more computation time than those with single rooms.

\begin{table*}[!t]
\centering
\caption{Detailed Specifications for Various Scenes in terms of size, room configurations, and materials. EM-GANSim is able to predict the signal power strength at any given data location in a few milliseconds.}
\label{table:detailed}
\resizebox{\textwidth}{!}{%
\begin{tabular}{|l|c|c|c|c|c|}
\hline
\textbf{Scene\#} & \textbf{1} & \textbf{2} & \textbf{3} & \textbf{4} & \textbf{5} \\
\hline
\textbf{Type} & Multiple rooms & Multiple rooms & Multiple rooms & Single room & Complex floor plan \\
\hline
\textbf{Size ($m^2$)} & 25 & 25 & 25 & 144 & 144 \\
\hline
\textbf{Materials used} & wood, concrete & wood, concrete, glass & wood, concrete & concrete & wood, concrete, glass \\
\hline
\end{tabular}%
}

\vspace{0.5cm}

\resizebox{\textwidth}{!}{%
\begin{tabular}{|l|c|c|c|c|c|}
\hline
\textbf{Scene\#} & \textbf{6} & \textbf{7} & \textbf{8} & \textbf{9} & \textbf{10} \\
\hline
\textbf{Type} & Complex floor plan & Complex floor plan & Single room & Single room & Single room \\
\hline
\textbf{Size ($m^2$)} & 144 & 16 & 16 & 16 & 144 \\
\hline
\textbf{Materials used} & wood, concrete, glass & wood, concrete, glass & wood, concrete, glass & concrete, glass & concrete, glass \\
\hline
\end{tabular}%
}

\vspace{0.5cm}

\resizebox{\textwidth}{!}{%
\begin{tabular}{|l|c|c|c|c|c|}
\hline
\textbf{Scene\#} & \textbf{11} & \textbf{12} & \textbf{13} & \textbf{14} & \textbf{15} \\
\hline
\textbf{Type} & Complex floor plan & Multiple rooms & Single room & Single room & Multiple rooms \\
\hline
\textbf{Size ($m^2$)} & 144 & 144 & 4 & 16 & 64 \\
\hline
\textbf{Materials used} & wood, concrete, glass & wood, concrete, glass & concrete & concrete, glass & wood, concrete, glass \\
\hline
\end{tabular}%
}
\end{table*}

\subsection{Accuracy of our Approach}

\begin{figure}[htbp]
\centerline{\includegraphics[scale=0.37]{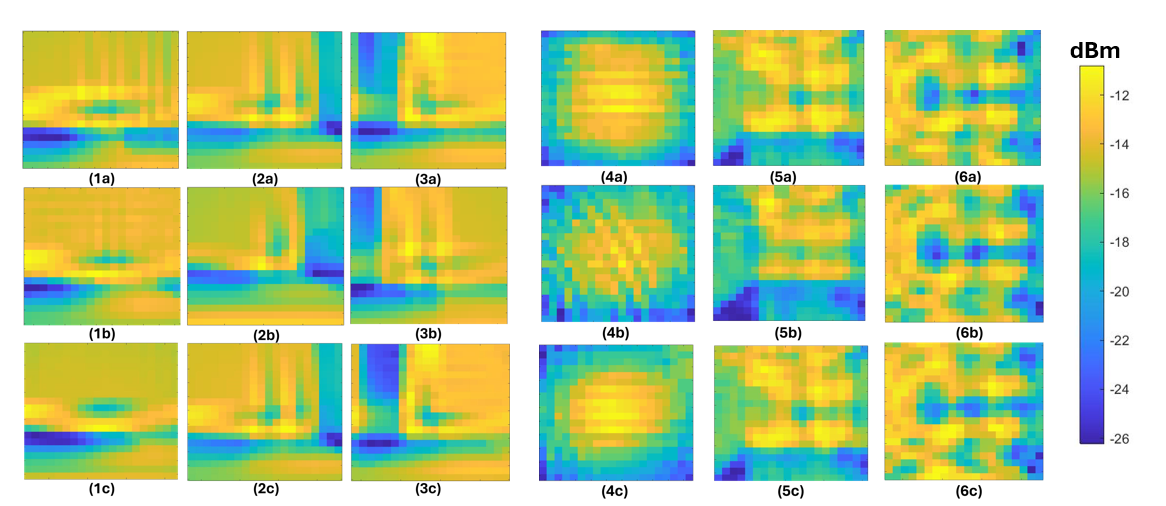}}
\caption{Heatmap comparisons for six indoor scenes (Scenes~1--6). Columns correspond to scenes~1--3 (left; 5\(\times\)5~m) and scenes~4--6 (right; 12\(\times\)12~m). Rows show (a) WinProp baseline, (b) GAN-based predictions, and (c) DCEM reference. Mean squared error values are reported in Table~\ref{table:MSE1}.}
\label{fig2}
\end{figure}

\begin{figure}[htbp]
\centerline{\includegraphics[scale=0.4]{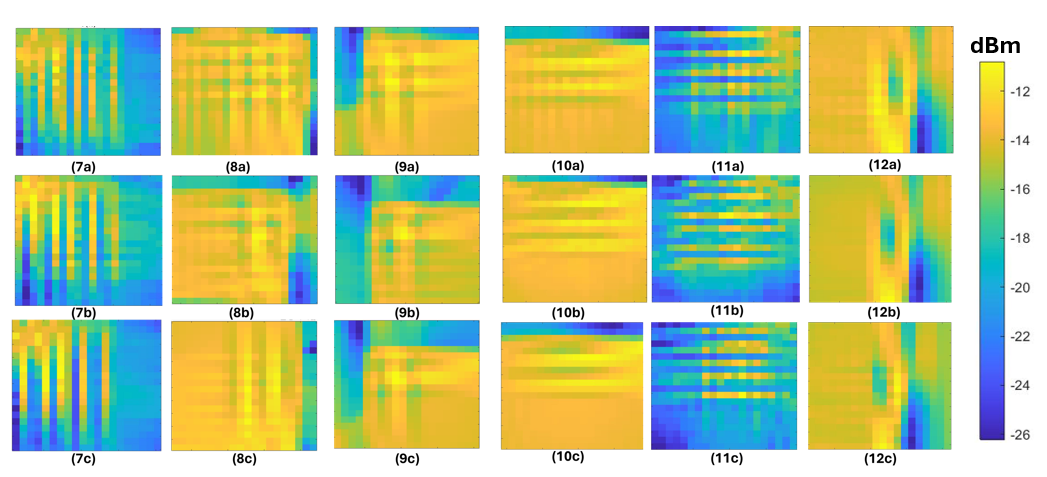}}
\caption{Heatmap comparisons for scenes~7--12. Columns correspond to scenes~7--9 (left) and scenes~10--12 (right), with larger rooms on the right. Rows show (a) WinProp baseline, (b) GAN-based predictions, and (c) DCEM reference.}
\label{fig6}
\end{figure}

The accuracy of our method is assessed by comparing the simulated received power distributions against standard RT simulations generated using DCEM~\cite{wang2022dynamic} and WinProp \citep{jakobus2018recent} simulators on the held-out benchmark scenes in Fig.~\ref{fig2} and Fig.~\ref{fig6}. We see that GAN-based tends to have a larger received power MSE than DCEM, which suggests some accuracy degradation while achieving the fastest running time among other methods. These heatmaps serve several purposes in supplementing the results: (1) Held-out Evaluation Across Diverse Conditions: These plots demonstrate the model's ability to generalize across different environments by presenting additional scenarios, validating its robustness and adaptability. (2) Comparative Analysis: The plots include comparisons between the EM-GANSim model predictions and those from benchmark WinProp. This comparative analysis highlights the strengths of EM-GANSim in terms of accuracy. (3) Visualization of EM Interactions: The heatmaps visually depict the power distribution and signal propagation across different room layouts. This visualization aids in understanding how well the model captures physical phenomena such as reflection and diffraction.

The comparison shows that EM-GANSim preserves the main spatial structure of the received-power maps while achieving substantially faster inference than full ray tracing. Unlike conventional machine learning baselines that focus on point-wise prediction, our approach preserves the global spatial structure of electromagnetic fields, which is critical for downstream wireless analysis and planning tasks. Although some accuracy degradation is observed relative to DCEM, the generated heatmaps remain broadly consistent with both DCEM and WinProp across diverse indoor scenes. 

\begin{table}[h]
\centering
\caption{MSE of GAN-based and DCEM compared to WinProp}
\label{table:MSE1}
\begin{tabular}{|l|c|c|}
\hline
\textbf{} & \textbf{GAN-based ($dbm^2$)} & \textbf{DCEM($dbm^2$)} \\
\hline
\textbf{Scene 1} & 7.29 & 5.60 \\
\textbf{Scene 2} & 9.47 & 9.08 \\
\textbf{Scene 3} & 8.51 & 11.00  \\
\textbf{Scene 4} & 12.03 & 6.42 \\
\textbf{Scene 5} & 11.71 & 9.44 \\
\textbf{Scene 6} & 5.91 & 7.36 \\
\textbf{Scene 7} & 7.66 & 10.93 \\
\textbf{Scene 8} & 7.93 & 4.47 \\
\textbf{Scene 9} & 9.76 & 7.95\\
\textbf{Scene 10} & 8.35 & 6.72 \\
\textbf{Scene 11} & 8.67 & 8.61 \\
\textbf{Scene 12} & 6.94 & 7.12\\
\hline
\end{tabular}
\end{table}

These benchmark scenes are held out from both the training and validation sets. The average MSE of GAN-based results of the training set is approximately 3 $dbm^2$ and that of the testing set is around 8.5 $dbm^2$.

\begin{figure}[htbp]
\centerline{\includegraphics[scale=0.48]{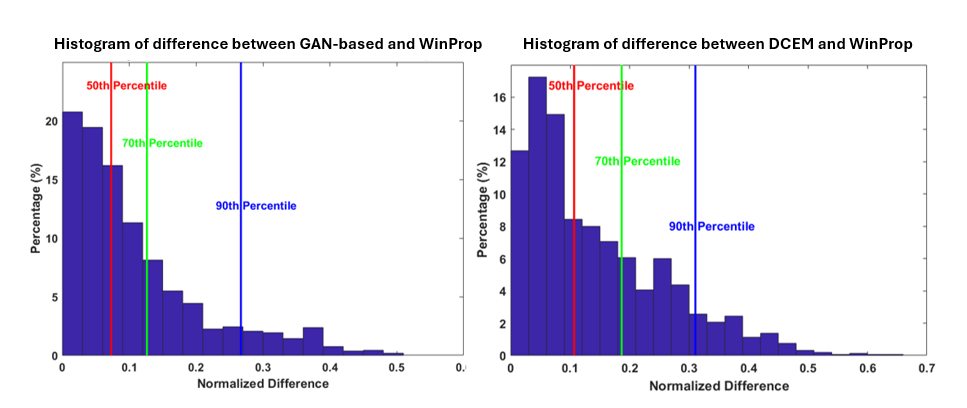}}
\caption{\textbf{Left Histogram:} Distribution of the normalized differences in received power levels between the GAN-based simulation and the WinProp simulation for the third-row scene. The vertical lines represent the 50th (median), 70th, and 90th percentiles, indicating a central tendency and spread of the differences. \textbf{Right Histogram:} Distribution of the normalized differences in received power levels between the DCEM simulation and WinProp simulation for the third-row scene. The percentiles are marked similarly. The left histogram appears slightly more concentrated in this example, suggesting that the GAN-based result is competitive with WinProp for this scene. However, the full quantitative comparison across all scenes is reported in Table III.}
\label{fig3}
\end{figure}

In Fig.~\ref{fig3}, we show a histogram distribution comparison of the normalized difference in the scenes in the third row of Fig.~\ref{fig2} (Scene 3). 

Our qualitative results (e.g., Figs.~\ref{fig2}–\ref{fig9}) reveal distinct lobes corresponding to line‑of‑sight, reflected, and diffracted contributions, consistent with the theoretical electromagnetic analyses in \citep{b24,b25,b26}. Although our network outputs a single heatmap rather than separate channels for each path type, the physics-based losses used during training encourage the generator to learn the aggregate superposition of these components. Hence the bright regions adjacent to walls and the smooth transitions at corners in our predicted heatmaps qualitatively align with specular reflection and diffraction reported in classical UTD literature.

\subsection{Efficiency Improvement through GAN}

To evaluate the efficiency of using GAN for quick simulations, the computation time was measured and compared between the GAN-based method and the traditional RT approach. The third column in Table ~\ref{table:computation_time} is the average generation time of data points in GAN-based predictions, calculated from total time divided by the total number of simulated points. For instance, in a single room of $2m * 2m$, with a resolution of $0.05m$, there are 1600 generated data points. Thus, each data point is generated in approximately 2 milliseconds.
\begin{table*}[h]
\centering
\caption{Computation Time Comparison Between GAN-based and Traditional RT Approaches}
\label{table:computation_time}
\resizebox{\textwidth}{!}{%
\begin{tabular}{|l|c|c|c|}
\hline
\textbf{} & \textbf{GAN-based (seconds)} & \textbf{Traditional RT (seconds)} & \textbf{Generation time per data point (seconds)} \\
\hline
\textbf{Single room (\textasciitilde 2*2 $m^2$)} & 3.2 & 12 & 0.002 \\
\textbf{Multiple rooms (\textasciitilde 8*8 $m^2$)} & 3 & 20 & 0.001 \\
\textbf{Complex floor plan (\textasciitilde 12*12 $m^2$)} & 4.3 & 22 & 0.0009 \\
\hline
\end{tabular}%
}
\end{table*}

The GAN-based method demonstrated a substantial reduction in computation time, offering near-instantaneous simulation results. This efficiency makes the GAN-based approach particularly suitable for applications requiring real-time data analysis and decision-making.

This dual-simulator evaluation separates training supervision from independent validation. DCEM provides the reference heatmaps and propagation masks used by the proposed losses, whereas WinProp provides an external ray-tracing baseline. Agreement with both simulators gives a stronger indication that the generated heatmaps are not merely reproducing simulator-specific artifacts.

Based on the comparisons after GAN training, we highlight the benefits of GAN below:
\begin{itemize}
    \item \textbf{High-Quality Synthetic Data Generation:} cGANs are adept at generating synthetic data that closely mirrors the distribution of real data, an essential capability for accurately predicting heatmaps from limited real-world data.
    
    \item \textbf{Efficiency in Prediction:} The GAN-based method can predict heat maps for an entire target area in a single inference step, offering a significant efficiency advantage over traditional, computation-intensive methods.
    
    \item \textbf{Accuracy Close to Ray Tracing Simulations:} cGANs have the potential to achieve accuracy levels comparable to those of traditional ray tracing simulations by learning to capture the complex variability of path loss across different environments.
\end{itemize}

\begin{figure}[htbp]
\centerline{\includegraphics[scale=0.35]{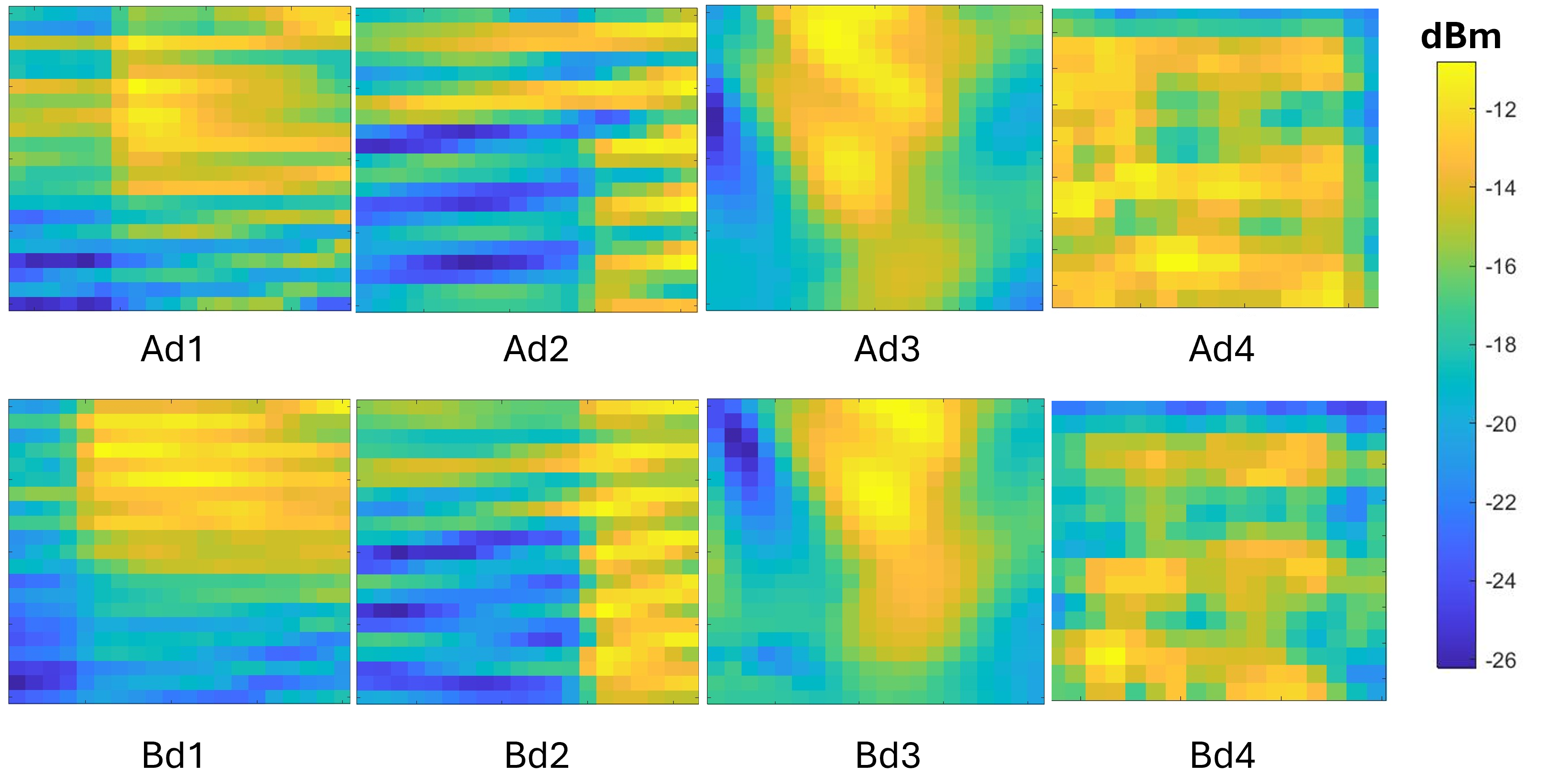}}
\caption{Additional evaluation on four new indoor scenes (Ad1--Ad4) used only for qualitative generalization analysis. Rows show (a) WinProp baseline and (b) GAN-based predictions. These examples illustrate generalization beyond the fifteen baseline benchmark scenes summarized in Table~\ref{table:detailed} and Figs.~\ref{fig2}--\ref{fig8}.}
\label{fig9}
\end{figure}

Fig.~\ref{fig9} aims to show the robustness and generalization of our EM-GANSim approach across diverse conditions. The CAD models used to generate these plots are derived from a dataset of 3D indoor environments, which is discussed in Section III. These models are selected to reflect the complexity and diversity of real-world indoor environments. This complexity arises from several factors: (1) Varied Room Configurations: The models include multiple room layouts with different sizes and shapes, ranging from simple square rooms to intricate floor plans with interconnected spaces and corridors. (2) Material Diversity: The inclusion of diverse materials like concrete, wood, and glass helps simulate the varying reflective, absorptive, and diffractive properties found in actual buildings. (3) Obstacles and Furnishings: The models feature obstacles such as walls and partitions, which affect EM wave propagation through reflection, diffraction, and scattering. The first row demonstrates results from a benchmark method (WinProp) for comparison. The second row of plots represents predictions from the EM-GANSim model, showcasing its capability to accurately predict electromagnetic wave interactions in various indoor environments. Fig.~\ref{fig9} also supports: (1) Validation Across Diverse Conditions: The plots demonstrate the model's ability to generalize across different environments by presenting additional scenarios, validating its robustness and adaptability. (2) Comparative Analysis: The plots include comparisons between the EM-GANSim model predictions and those from benchmark WinProp. This comparative analysis highlights the strengths of EM-GANSim in terms of accuracy. (3) Visualization of EM Interactions: The heatmaps visually depict the power distribution and signal propagation across different room layouts. This visualization aids in understanding how well the model captures physical phenomena such as reflection and diffraction. 

Overall, the results show that EM-GANSim achieves substantially lower inference time than ray tracing while maintaining received-power maps that remain close to both DCEM and WinProp across diverse scenes. As expected, some accuracy degradation is observed relative to full ray tracing, but the trade-off is favorable for real-time simulation settings.

\section{Ablation Experiments}
\begin{figure}[htbp]
\centerline{\includegraphics[scale=0.55]{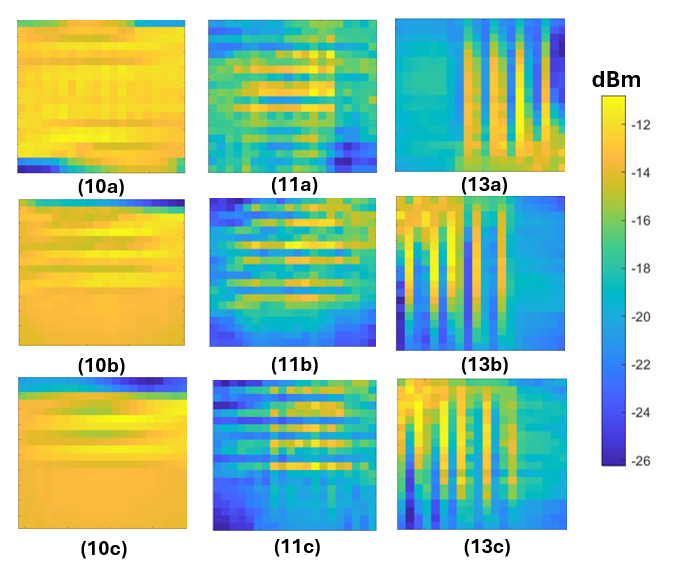}}
\caption{Comparative heatmaps of received power distributions across scenes~10,~11, and~13: (a) predictions from a GAN trained without noise input; (b) predictions from a GAN trained with noise input; and (c) DCEM benchmark results.}
\label{fig7}
\end{figure}

 In this section, we first analyze the effect of excluding Gaussian noise from the training process, an element typically introduced to break the symmetry in the model weights, ensuring that different units learn different features.  We verify the noise's impact on the generated heatmaps and their respective MSEs. By comparing heatmaps and MSE values, we evaluate the GAN's performance in generating received power distribution in the absence of noise. This ablation study serves not only to reinforce the validity of our methodology but also to offer insights that could refine future implementations of machine learning in EM ray tracing. We define the ablation experiment results as GAN-No-Noise.
Observations based on the heatmaps in Fig.~\ref{fig7}, from the GAN-No-Noise, GAN-based, and DCEM predictions in testing scenes are discussed as follows:
\begin{itemize}
\item \textbf{First Row (GAN-No-Noise):} The absence of Gaussian noise results in less varied and uniform heatmaps, indicating potential over-smoothing and reduced accuracy in capturing EM wave interactions within the environment.

\item \textbf{Second Row (GAN-based with Noise):} Inclusion of noise introduces more defined contrasts and a broader range of power levels, suggesting a better representation of the complex nature of EM propagation and environmental features.

\item \textbf{Cons of GAN-No-Noise:} Lack of noise in training leads to simpler patterns, reduced model accuracy, and potential issues in generalizing to new environments, which is critical for applications like network planning.

\item \textbf{Importance of Noise:} Gaussian noise is essential in training to break symmetry in the model, ensuring diverse learning and preventing the network from collapsing into repetitive pattern production.
\end{itemize}

These observations underline the importance of including noise in the GAN training process to enhance the model's ability to predict received power distributions accurately and robustly, especially when applied to complex indoor EM propagation scenarios. Gaussian noise plays a crucial role in breaking weight symmetry during training and promoting diversity in generator outputs. Removing such noise results in less varied, overly smoothed heatmaps, reflecting reduced accuracy and an increased likelihood of mode collapse.

We also include the corresponding MSE of GAN-No-Noise and GAN-based compared to DCEM in Table \ref{table:MSE2} below. For these three testing cases, GAN-based predictions consistently have lower MSE values than GAN-No-Noise, indicating that the inclusion of noise during the training process contributes to a more accurate prediction of received power levels. The improved MSE with noise suggests that Gaussian noise acts as a regularizer, preventing the model from memorizing the training data and instead forcing it to learn the underlying distribution. The presence of noise also introduces a wider variety of scenarios during training, making the GAN model more robust to unseen environments and better at generalizing from the training data.

\begin{table}[h]
\centering
\caption{MSE of GAN-No-Noise and GAN-Based compared to DCEM}
\label{table:MSE2}
\begin{tabular}{|l|c|c|}
\hline
\textbf{} & \textbf{GAN-No-Noise ($dbm^2$)} & \textbf{GAN-based($dbm^2$)} \\
\hline
\textbf{Scene 10} & 10.88 & 5.24\\
\textbf{Scene 11} & 9.52 & 7.19 \\
\textbf{Scene 13} & 16.38 & 3.65 \\

\hline
\end{tabular}
\end{table}

Another ablation test is designed to evaluate the impact of incorporating physical constraints into the objective function of our generator. These physical constraints are integrated to ensure that the generated samples adhere to the fundamental principles of electromagnetic wave propagation, accounting for direct propagation, reflection, and diffraction effects. The ablation tests will involve running the simulator under two distinct conditions: 
\begin{itemize}
    \item \textbf{With Physics Constraints}: The generator's objective function will include the physical constraints loss ($\mathcal{L}_{\text{phy}}$), which comprises terms for direct path propagation ($\mathcal{L}_{\text{direct}}$), reflections ($\mathcal{L}_{\text{ref}}$), and diffractions ($\mathcal{L}_{\text{diff}}$).
    \item \textbf{Without Physics Constraints}: The physical constraints loss ($\mathcal{L}_{\text{phy}}$) will be omitted from the objective function, leaving only the cGAN loss and the MSE loss components.
\end{itemize}

\begin{figure}[htbp]
\centerline{\includegraphics[scale=0.45]{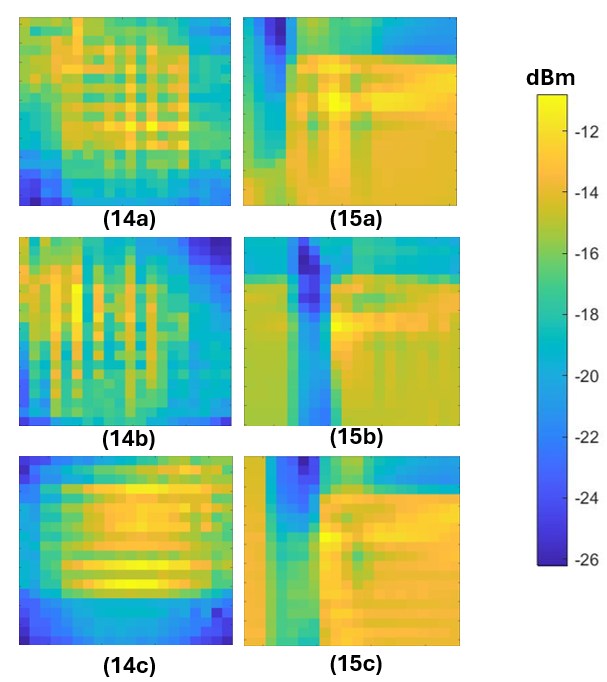}}
\caption{Comparative heatmaps for Scenes~14 and~15: (a) DCEM benchmark results; (b) GAN predictions trained with physics constraints; and (c) GAN predictions trained without physics constraints. The physics‑constrained predictions closely match the benchmark, whereas omitting constraints results in noticeable deviations.}
\label{fig8}
\end{figure}

This ablation test demonstrates the impact of incorporating physical constraints by comparing the performance and accuracy of the generator under both conditions as shown in Fig.~\ref{fig8}. This comparison highlights the crucial role of physical constraints in enhancing the accuracy and realism of the GAN-based model for simulating indoor signal propagation, as evidenced by the closer alignment with the DCEM benchmark. Key performance metrics observed were:

\begin{itemize}
    \item \textbf{Signal Propagation Accuracy}: The tests showed that the generator with physical constraints better preserves the spatial structure of the DCEM reference heatmaps. The improvement is reflected in masked received-power responses over direct, reflection, and diffraction regions, rather than in separate predicted ray powers. This supports the role of the physics-guided losses in capturing physically meaningful propagation patterns in indoor environments.
    \item \textbf{Visual and Structural Fidelity}: The generated samples with physical constraints exhibited higher visual realism and structural coherence. These samples were more accurate in modeling the indoor environments compared to those generated without the constraints.
\end{itemize}

\section{Conclusions, Limitations, and Future Work}
We present a novel approach that uses ML methods along with EM ray tracing to enhance the accuracy and efficiency of wireless communication simulation within 3D indoor environments. We use a modified cGAN that utilizes encoded geometry and transmitter location and can be used for accurate EM wave propagation. We have evaluated its performance on a large number of complex 3D indoor scenes and its performance is comparable to EM ray tracing-based simulations. Furthermore, we observe a 5X performance improvement over prior methods.

Our study enhances wireless communication efficiency and lays the ground for future real-time applications. Our approach has some limitations. Since our training data is based on ray tracing, our prediction scheme may not be able to accurately model low-frequency or other wave interactions. Our current approach is limited to indoor scenes, and we would also like to evaluate it in scenes with multiple dynamic objects. A key challenge is to extend and use these methods for large urban scenes with complex traffic patterns to model wireless signals.

Furthermore, we plan to add dynamic elements, such as movable partitions and furniture, to simulate real-world changes in indoor layouts. By leveraging publicly available architectural data (such as the 3D-Front dataset  \citep{fu20213d}), we will continuously update the dataset with new scenarios that reflect emerging trends in building design and technology. This comprehensive dataset expansion will improve the model’s ability to predict EM wave propagation in complex and varied indoor environments, ultimately enhancing its applicability and reliability in practical applications.

However, heterogeneous material types remain a significant challenge for our model because the current lookup-table encoding captures discrete material categories but does not explicitly model continuous electromagnetic parameters (e.g., permittivity or conductivity), which may limit generalization across diverse material classes. Future work could explore learning continuous material embeddings from spectral or multimodal inputs, or integrating additional modalities (e.g., optical or infrared imagery) to more accurately characterize material‑specific propagation effects and improve generalization across different building materials.

\begin{IEEEbiography}
[{\includegraphics[width=1in,height=1.25in,clip,keepaspectratio]{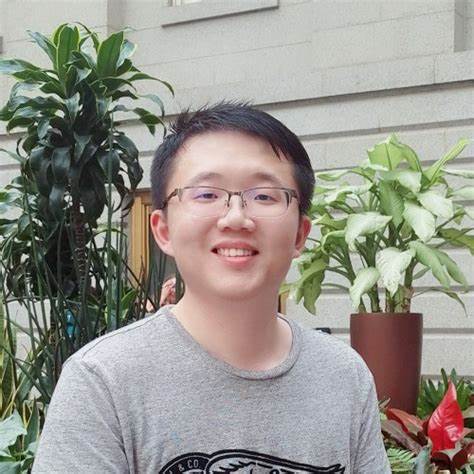}}]{Ruichen Wang} received the B.S. degree in astrophysics from Peking University in 2015 and the M.S. degree in electrical engineering from New York University in 2017. He recently completed the Ph.D. degree in electrical engineering from the University of Maryland, focusing on developing advanced propagation models and signal processing techniques for complex and dynamic indoor/outdoor environments.

Ruichen has authored or co-authored several papers in wireless communications and possesses experience in both academic and industry environments. His research contributions include developing a GAN-based approach for indoor signal simulations, creating novel algorithms for electromagnetic ray-tracing simulations, and enhancing predictive models in dynamic urban and complex indoor scenarios. His research interests encompass channel modeling, signal processing, and the application of machine learning techniques in communication systems.
\end{IEEEbiography}

\vspace{-40em}
\begin{IEEEbiography}[{\includegraphics[width=1in,height=1.25in,clip,keepaspectratio]{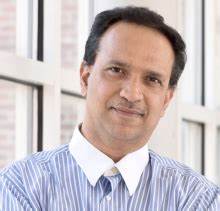}}]{Dinesh Manocha} is Paul Chrisman-Iribe Professor and a Distinguished University Professor of CS and ECE at the University of Maryland College Park. He has published more than 800+ papers and supervised 53+ PhD dissertations. He is an inventor of 17 patents, some of which licensed to industry.  His group has developed many widely used software systems (with 2M+ downloads) and licensed them to more than 60 commercial vendors. He is a Fellow of AAAI, AAAS, ACM, IEEE, NAI, Washington Academy of Sciences, member of ACM SIGGRAPH and IEEE VR Academies, and a recipient of the Bézier Award and Jimmy H. C. Lin Award. He received the Distinguished Alumni Award from IIT Delhi. He has co-founded multiple companies, including Impulsonic, which was acquired by Valve.
\end{IEEEbiography}

\end{document}